\documentclass[letterpaper]{article} 
\usepackage{aaai2026}
\usepackage[switch]{lineno}
\usepackage{amsmath}
\usepackage{times}  
\usepackage{helvet}  
\usepackage{courier}  
\usepackage[hyphens]{url}  
\usepackage{graphicx} 
\usepackage{natbib}  
\usepackage{caption} 
\usepackage{algorithm} 
\usepackage{algpseudocode}
\usepackage{standalone}

\usepackage{booktabs} 
\usepackage{multirow} 
\usepackage{siunitx}
\usepackage{tabularx}
\usepackage{array}
\usepackage[table]{xcolor}
\newcolumntype{C}{>{\centering\arraybackslash}X}
\newcolumntype{R}{>{\raggedleft\arraybackslash}X}
\usepackage{ragged2e}

\usepackage{xurl}

\usepackage{newfloat}
\usepackage{listings}
\usepackage{pifont}
\usepackage{enumitem}
\setenumerate[1]{itemsep=0pt,partopsep=0pt,parsep=\parskip,topsep=2pt}
\setitemize[1]{itemsep=0pt,partopsep=0pt,parsep=\parskip,topsep=2pt}
\setdescription{itemsep=0pt,partopsep=0pt,parsep=\parskip,topsep=2pt}

\DeclareCaptionStyle{ruled}{labelfont=normalfont,labelsep=colon,strut=off} 
\floatstyle{ruled}
\newfloat{listing}{tb}{lst}{}
\floatname{listing}{Listing}
\usepackage[textwidth=2cm, textsize=small]{todonotes} 

\title{From Query to Logic: Ontology-Driven Multi-Hop Reasoning in LLMs}

\author{
    Haonan Bian\textsuperscript{\rm 1},
    Yutao Qi\footnote{The corresponding author.}\textsuperscript{\rm 1},
    Rui Yang\textsuperscript{\rm 1},
    Yuanxi Che\textsuperscript{\rm 1},
    Jiaqian Wang\textsuperscript{\rm 1},
    Heming Xia\textsuperscript{\rm 2},
    Ranran Zhen\textsuperscript{\rm 3}
}
\affiliations{
    \textsuperscript{1}School of Cyber Security, Xidian University\\
    \textsuperscript{2}Department of Computing, The Hong Kong Polytechnic University\\
    \textsuperscript{3}School of Future Science and Engineering, Soochow University\\
    23151214251@stu.xidian.edu.cn, he-ming.xia@connect.polyu.hk, zenrran@gmail.com
}

\usepackage{bibentry}

\begin{document}

\maketitle

\begin{abstract}
Large Language Models (LLMs), despite their success in question answering, exhibit limitations in complex multi-hop question answering (MQA) tasks that necessitate non-linear, structured reasoning. 
This limitation stems from their inability to adequately capture deep conceptual relationships between entities.
To overcome this challenge, we present \textbf{ORACLE} (\textbf{O}ntology-driven \textbf{R}easoning \textbf{A}nd \textbf{C}hain for \textbf{L}ogical \textbf{E}lucidation), a training-free framework that combines LLMs' generative capabilities with the structural benefits of knowledge graphs.
Our approach operates through three stages: (1) dynamic construction of question-specific knowledge ontologies using LLMs, (2) transformation of these ontologies into First-Order Logic (FOL) reasoning chains, and (3) systematic decomposition of the original query into logically coherent sub-questions.
Experimental results on multiple standard MQA benchmarks show that our framework achieves highly competitive performance, rivaling current state-of-the-art models like DeepSeek-R1. 
Detailed analyses further confirm the effectiveness of each component, while demonstrating that our method generates more logical and interpretable reasoning chains than existing approaches.
\end{abstract}


\section{Introduction}
Large Language Models have demonstrated significant success in knowledge-based question answering~\cite{brown2020language, deepseekai2025deepseekr1incentivizingreasoningcapability, qwen3}. However, they continue to face challenges in MQA tasks~\cite{wang_learning_2024,iter_rag}, which require integrating and reasoning over multiple, discrete sources of information. 
A significant challenge, as highlighted in \citet{ju-etal-2024-investigating}, is that LLMs tend to rely on guessing derived from training data rather than actually reasoning in MQA tasks. Therefore, the key lies in advancing beyond the generation of factually correct chains toward empowering LLMs to uncover and leverage deeper conceptual relationships between entities—a capability essential for true multi-hop understanding.

Recent research on MQA has predominantly focused on two paradigms: prompting strategies and Retrieval-Augmented Generation (RAG). 
Standard RAG approaches often falter in MQA as they struggle to retrieve all necessary information fragments in a single pass. 
Iterative retrieval methods like ReAct~\cite{yao_react_2023} and EfficientRAG~\cite{zhuang2024efficientrag} address this by progressively refining the retrieved results. Specifically, EfficientRAG employs a smaller model as an evaluator and retrieval generator to streamline the process without constant reliance on a large model. 
Another line of methods, exemplified by PAR RAG~\cite{zhang2025credible}, centers on upfront problem decomposition to facilitate more globally coherent solution planning.
Concurrently, researchers have proposed leveraging Knowledge Graphs (KGs) to represent the MQA reasoning process, capitalizing on their structured nature. 
Among these studies, LPKG~\cite{wang_learning_2024} utilizes inherent patterns within a KG to guide the LLM's decomposition and planning, while ROG~\cite{luo_reasoning_2024} uses reasoning paths from KG subgraphs to direct retrieval and problem-solving.

While these KG-based methods show promise, we argue that relying on predefined structural paths is insufficient. 
We posit that the essence of complex reasoning lies not only in the relationships between entities but also in the ``concepts" they belong to and the hierarchical relationships between these concepts. 
To this end, we introduce \textbf{ORACLE}, a training-free MQA framework centered on a \textit{Question-centric Knowledge Graph Ontology}. 
Instead of using static KG paths or requiring model fine-tuning, our framework dynamically constructs a bespoke ontology for each question using a powerful LLM. 
This ontology provides a structured semantic scaffold, capturing the core entities, their interrelations, and underlying conceptual hierarchies, thereby guiding the LLM's reasoning process.

ORACLE operates in three sequential stages: (1) ontology construction, (2) First-Order Logic (FOL) formulation, and (3) sub-question decomposition. 
First, the LLM dynamically constructs a question-centric knowledge ontology. 
This ontology delineates the key entities within the question and their underlying conceptual relationships, establishing a structured foundation for the reasoning process. 
Subsequently, this ontology facilitates the translation of the original question into a formal FOL reasoning chain. 
This logical structure makes the required inferential steps explicit. In the final stage, guided by both the knowledge ontology and the FOL chain, the LLM systematically decomposes the complex initial query into a sequence of simpler, logically coherent sub-questions. 

To sum up, our key contributions are as follows:
\begin{itemize}
\item To the best of our knowledge, this is the first work to apply ontology theory to guide LLM reasoning by introducing a dynamic \textit{Question-centric Knowledge Ontology} to enhance sub-problem decomposition in MQA.
\item We propose ORACLE, a training-free MQA framework that integrates ontological reasoning with FOL, achieving competitive performance on standard benchmarks.
\item Our analysis demonstrates that ORACLE generates more interpretable reasoning chains than path-pattern approaches, providing new insights into LLM MQA reasoning.
\end{itemize}

\section{Related Work}
\paragraph{Multi-Hop Question Answering}
Before the era of Large Language Models, the research field of multi-hop question answering (MQA) was dominated by Graph Neural Network approaches that modeled entity relationships and dynamic reasoning processes~\cite{fang-etal-2020-hierarchical, zhang2022edge, li2022question}; however, these methods were fundamentally limited by their reliance on dataset-specific training, hindering their generalizability.
The advent of LLMs shifted the MQA paradigm towards in-context learning, notably through Chain-of-Thought (CoT) prompting~\cite{wei_chain--thought_2023}. To address the limitations of LLMs, such as static knowledge and hallucination, Retrieval-Augmented Generation (RAG) has emerged as a key framework, spurring two main research directions. One line of inquiry focuses on refining RAG, with methods like ReAct~\cite{yao_react_2023} introducing iterative retrieval and EfficientRAG~\cite{zhuang2024efficientrag} using a smaller model for query refinement. Another prominent direction is question decomposition, where approaches like PAR-RAG~\cite{zhang2025credible} break down complex questions into sub-plans, and LPKG~\cite{wang_learning_2024} learns decomposition strategies from knowledge graphs (KGs). Our work advances this research on question decomposition. Instead of learning patterns from a static KG, we leverage knowledge representation principles to dynamically generate an ontology that structures the question into a logical chain.

\paragraph{KG-enhanced LLMs}
Knowledge Graphs (KGs) are widely used to provide structured, factual grounding for LLMs. One prominent line of research integrates KGs directly into the LLM inference process, for instance by augmenting context with retrieved triples (TOG~\cite{sun_think--graph_2023}), generating relation-aware plans to mitigate hallucination (ROG~\cite{luo_reasoning_2024}), or constraining the decoding process with KG paths to ensure faithful reasoning (GCR~\cite{luo_graph-constrained_2024}). Another direction utilizes KGs to synthesize high-quality reasoning data for model fine-tuning, as demonstrated by MedReason~\cite{wu_medreason_2025} and OntoTune~\cite{liu_ontotune_2025}.
While our work is conceptually aligned with methods like LPKG~\cite{wang_learning_2024} that use KG structures for task decomposition, our key distinction is the dynamic use of KG schema. Rather than relying on static KG patterns, our framework constructs a question-centric ontology on-the-fly to guide the LLM's reasoning process.

\paragraph{First-Order Logic (FOL)}
Recent work integrating LLMs with symbolic reasoning, such as FOL, has pursued two primary strategies~\cite{ye_generating_2023, gaur_reasoning_2023}. The first uses LLM to translate natural language into formal logic, which is then processed by an external symbolic reasoner for accurate inference~\cite{pan_logic-lm_2023, olausson_linc_2023}. The second strategy aims to enhance the LLM's intrinsic reasoning capabilities, either through symbolic chain-of-thought prompting~\cite{xu_faithful_2024} or fine-tuning on logic-based datasets~\cite{morishita_enhancing_2024, morishita_learning_nodate}. However, these approaches are limited by their reliance on formal logic, which often fails to capture the rich context of real-world relationships.
Our work addresses this gap by synergizing formal logical rules with structured knowledge from knowledge graphs. This enables the LLM to handle complex questions requiring both formal deduction and real-world contextual understanding.

\begin{figure*}[h]
    \centering
    \includegraphics[width=0.95\linewidth]{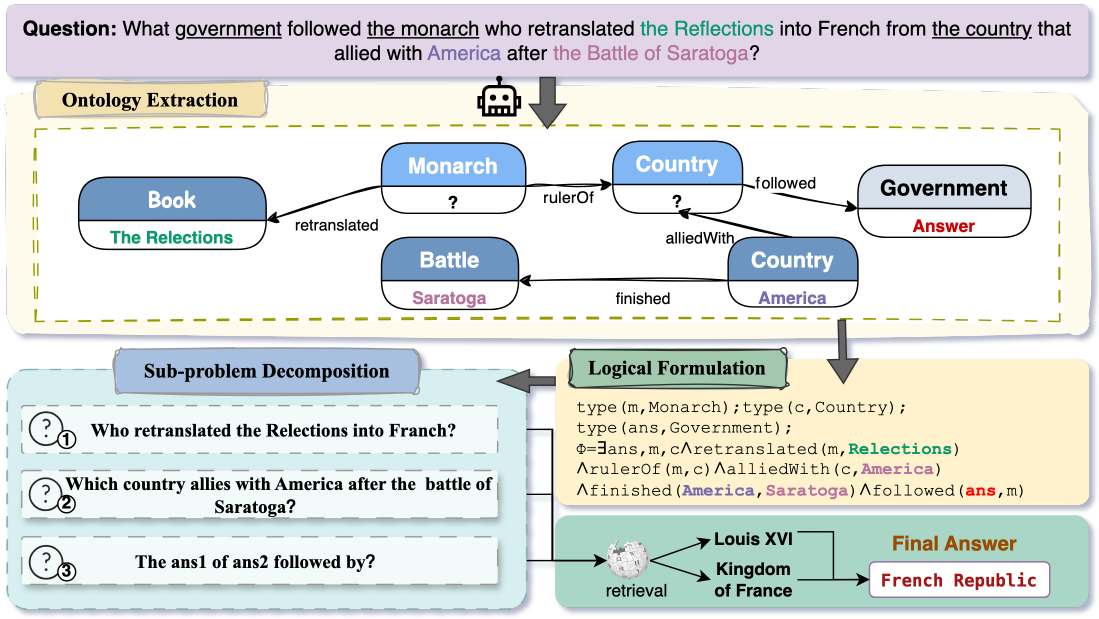}
\caption{Overview of Our Proposed ORACLE Framework, Which Consists of Three Modules: (1) Dynamic Construction of Question-Specific Knowledge Ontologies Using LLMs, (2) Transformation of These Ontologies into FOL Reasoning Chains, and (3) Systematic Decomposition of the Original Query into Logically Coherent Sub-Questions.}
    \label{fig:method_framework}
\end{figure*}

\section{Preliminary}
This section establishes the core terminology and notation used in this paper. An ontology provides the formal schema for a KG, defining the concepts, attributes, and semantic relationships within a domain to enable structured reasoning. 
\begin{itemize}
    \item \textbf{KG ($G$)}: A KG is a structured representation of factual knowledge. Formally, it is a directed graph defined by a set of triples, $G \subseteq E \times R \times E$.
    \item \textbf{Entity ($E$)}: The set of nodes in the graph, where each entity $e \in E$ represents a real-world object or abstract concept. In a triple, $h$ and $t$ denote the head and tail entities, respectively.
    \item \textbf{Relation ($R$)}: The set of directed, typed edges, where each relation $r \in R$ represents a specific type of link between entities.
    \item \textbf{Ontology ($O$)}: A formal specification that defines the schema of a KG. It includes the vocabulary of classes and relations, their properties, and the constraints that govern their structure.
    \item \textbf{Class ($C$)}: The set of categories that group entities. Each class $c \in C$ represents a collection of entities sharing common characteristics (e.g., Person, Location).
    \item \textbf{Instance of}: A predicate, denoted as $\text{type}(e, c)$, which asserts that an entity $e \in E$ is an instance of a particular class $c \in C$.
    \item \textbf{Prompt Concatenation ($\oplus$)}: An operator used to denote the sequential concatenation of prompts or text segments, i.e., $P_1 \oplus P_2$ appends $P_2$ after $P_1$ with a delimiter (such as a newline or separator token).
\end{itemize}

\section{Methodology}
Our framework for MQA (Figure~\ref{fig:method_framework}) has three stages: \textit{Ontology Extraction}, \textit{FOL Construction}, and \textit{Sub-question Decomposition}. First, an LLM extracts a question-specific ontology from the query. This ontology is then used to construct a FOL formula. Finally, the FOL formula decomposes the complex query into simpler sub-questions, which are executed to derive the final answer.

\subsection{Ontology Extraction}
The initial stage, Ontology Extraction, aims to construct a lightweight, question-specific ontology, denoted as $O_Q$, from the natural language question $Q$. This process is driven by an LLM that functions as a knowledge extractor. Unlike traditional methods that focus solely on identifying entities ($E$) and relations ($R$), our approach emphasizes the extraction of the underlying entity classes ($\mathcal{C}$). This class-centric ontology, $O_Q$, serves as a critical semantic and structural guide for subsequent reasoning steps. To further improve the model's planning, we also instruct the LLM to predict the class of the final answer. This extraction process is formally represented as:
\begin{equation*}
O_Q = (C_Q, R_Q) = f_{\text{LLM}}(Q)
\end{equation*}
Here, $C_Q \subseteq \mathcal{C}$ is the set of relevant classes and their corresponding entities, and $R_Q \subseteq R$ is the set of relations that guides the decomposition process.

For instance, as illustrated in Figure~\ref{fig:method_framework}, 
given the question $Q$ = \textit{“What government followed the monarch who retranslated the Reflections into French from the country that allied with America after the Battle of Saratoga?”}, 
this stage extracts:  

\begin{itemize}
    \item \textbf{Classes and Entities ($C_Q$):} 
    \begin{equation*}
    \begin{aligned}
       & \text{type}(\text{The Reflections}, Book), \\
       & \text{type}(m, Monarch), \\
       & \text{type}(c, Country), \\
       & \text{type}(Saratoga, Battle), \\
       & \text{type}(ans, Government)
    \end{aligned}
    \end{equation*}
    where $m$, $c$, and $ans$ are variables. 
    Notably, $ans$ denotes the final answer entity (the Government).
    \item \textbf{Relations ($R_Q$):} 
      $m \xrightarrow{\texttt{retranslated}}$ \textit{The Reflections};  
      $m \xrightarrow{\texttt{rulerOf}} c$;  
      $c \xrightarrow{\texttt{followed}} ans$;  
      $c \xrightarrow{\texttt{alliedWith}}$ \textit{America};  
      \textit{Saratoga} $\xrightarrow{\texttt{finished}}$ \textit{America}.      
\end{itemize}

This structure corresponds to the ontology graph shown in Figure~\ref{fig:method_framework}, 
where nodes denote classes and instantiated entities, and directed edges denote relations among them.

\noindent
\paragraph{Why Ontology is Necessary.} 
While LLMs inherently possess background knowledge about many entities, direct decoding often relies on local token-level similarity, which can easily introduce irrelevant entities into the reasoning chain. For instance, as illustrated in our case study (Appendix~\ref{appendix:case-studies}), a baseline decomposition mistakenly aligns \emph{The Blonde from Singapore} with \emph{The Blonde from Peking}, a phenomenon we term \emph{logicalized semantic drift}.

To mitigate this issue, we introduce a lightweight ontology as an intermediate scaffold. The ontology enforces (i) \textbf{type constraints}, ensuring that variables are grounded to the correct conceptual classes, (ii) \textbf{relational consistency}, aligning predicates with the intended semantic hierarchy, and (iii) \textbf{path stability}, as demonstrated in the appendix case studies (Appendix~\ref{appendix:case-studies}). Together, these properties substantially reduce semantic drift and provide a global structural perspective before entering the FOL construction stage.

\subsection{FOL Construction}
In the FOL Construction stage, the framework converts the extracted ontology $O_Q$ into a precise and interpretable FOL formula, denoted as $\Phi$. This transformation methodically maps the ontology's components into a logical structure: relations $r \in R_Q$ become predicates, while entity classes $c \in C_Q$ enforce type constraints on the variables. The output is a formal logical expression of the original question $Q$, complete with existential quantifiers ($\exists$) and a designated answer variable.

As illustrated in Figure~\ref{fig:method_framework}, the question $Q$ is transformed into the following formula $\Phi$, accompanied by type declarations. The predicate names in the example are aligned with the ontology graph for clarity.

\textbf{Type Constraints:} 
\begin{equation*}
\begin{aligned}
   & \text{type}(m, Monarch), \\
   & \text{type}(c, Country), \\
   & \text{type}(ans, Government)
\end{aligned}
\end{equation*}

\textbf{FOL Formula:}
\begin{align*}
\Phi = & \exists ~ans, m, c \quad \\
& \land \text{retranslated}(m, \text{The Reflections}, French) \\
& \land \text{rulerOf}(m, c) \\
& \land \text{alliedWith}(c, America) \\
& \land \text{finished}(Saratoga, America) \\
& \land \text{followed}(c, ans)
\end{align*}

For instance, the clause $rulerOf(m, c)$ uses the predicate \texttt{rulerOf} to link the variables $m$ and $c$, which are constrained by the types \texttt{Monarch} and \texttt{Country}, respectively. This demonstrates how the logical formula directly mirrors the structure of the extracted ontology, thereby aiding in the decomposition of the question.

\subsection{Sub-question Decomposition}
The final stage, Sub-question Decomposition, breaks down the complex query into an ordered sequence of simpler, solvable sub-questions $\{Q_1, Q_2, \dots, Q_n\}$. To achieve this, we leverage the instruction-following capabilities of an LLM. The model is prompted with a comprehensive input that includes the original question $Q$, the extracted ontology $O_Q$, and the generated FOL formula $\Phi$. Guided by this rich context, particularly the logical structure of $\Phi$, the LLM formulates a multi-step reasoning plan. Each step in this plan materializes as a distinct natural language sub-question, $Q_i$. A key feature of this process is the use of placeholders to dynamically insert answers from preceding steps into subsequent questions, creating a coherent and executable reasoning chain. This decomposition is formally represented as:
\begin{equation*}
\{Q_1, Q_2, \dots, Q_n\} = g_{\text{LLM}}(Q, O_Q, \Phi)
\end{equation*}
These sub-questions are then executed sequentially to derive the final answer. For each step $i$, the sub-question $Q_i$ is first formulated into a subquestion $Q'_i$ by incorporating the set of necessary prior answers $A_{<i} = \{A_1, \dots, A_{i-1}\}$. A retriever, $\mathcal{R}$, then fetches relevant context based on this prompt, where $C_i = \mathcal{R}(Q'_i)$. Finally, an LLM generates the answer for the current step, $A_i = \mathcal{\text{LLM}}(Q'_i, C_i)$. This iterative process continues until the final sub-question, $Q_n$, is solved. Its answer, $A_n$, is returned as the final answer $\hat{y}$ to the original query. The overall algorithm of our proposed ORACLE framework is demonstrated in Algorithm~\ref{alg:framework_detailed}.

\begin{algorithm}[t]
\caption{Our proposed ORACLE framework}
\label{alg:framework_detailed}
\textbf{Input:} Input question $Q$, Large Language Model $\mathcal{LLM}$, Retriever $\mathcal{R}$, Ontology Prompt $\mathcal{P}_{Onto}$, FOL Prompt $\mathcal{P}_{FOL}$, Decomposition Prompt $\mathcal{P}_{Decomp}$ \\
\textbf{Output:} Final answer, $\hat{y}$
\begin{algorithmic}[1]

\State $O_Q \gets \mathcal{LLM}([\mathcal{P}_{Onto} \oplus Q])$
\State $\Phi \gets \mathcal{LLM}([\mathcal{P}_{FOL} \oplus O_Q])$
\State $\{Q_1, \dots, Q_n\} \gets \mathcal{LLM}([\mathcal{P}_{Decomp} \oplus Q \oplus O_Q \oplus \Phi])$

\State $A_{ans} \gets []$ 
\For{$i=1$ \textbf{to} $n$}
    \State $Q'_i \gets {substitute}(Q_i, A_{ans})$
    \State $ans_i \gets \mathcal{LLM}([\mathcal{R}(Q'_i) \oplus Q'_i])$
    \State Append $ans_i$ to $A_{ans}$
\EndFor

\State $\hat{y} \gets A_{ans}[n]$ \Comment{Return the final sub-answer}
\State \textbf{return} $\hat{y}$
\end{algorithmic}
\end{algorithm}

\begin{table*}[t]
\centering
\setlength{\tabcolsep}{1.4mm}

\begin{tabular}{llc rrrrrrrr}
\toprule
\multicolumn{1}{c}{\multirow{2}{*}{Method}} & \multicolumn{1}{c}{\multirow{2}{*}{Model}} & \multicolumn{1}{c}{\multirow{2}{*}{Planning}} & \multicolumn{2}{c}{HotPotQA} & \multicolumn{2}{c}{2WikiMQA} & \multicolumn{2}{c}{Musique} & \multicolumn{2}{c}{Average} \\
\cmidrule(lr){4-5} \cmidrule(lr){6-7} \cmidrule(lr){8-9} \cmidrule(lr){10-11}
\multicolumn{1}{c}{} & \multicolumn{1}{c}{} & \multicolumn{1}{c}{} & \multicolumn{1}{c}{EM} & \multicolumn{1}{c}{F1} & \multicolumn{1}{c}{EM} & \multicolumn{1}{c}{F1} & \multicolumn{1}{c}{EM} & \multicolumn{1}{c}{F1} & \multicolumn{1}{c}{EM} & \multicolumn{1}{c}{F1} \\
\midrule
NoCoT~\cite{ouyang2022training}               & gpt-3.5-turbo  & \ding{56}     & 0.306          & 0.429          & 0.271          & 0.316          & 0.058          & 0.162          & 0.212          & 0.302          \\
CoT~\cite{wei_chain--thought_2023}             & gpt-3.5-turbo  & \ding{56}     & 0.222          & 0.336          & 0.168          & 0.262          & 0.052          & 0.134          & 0.147          & 0.244          \\
RAG~\cite{gao2023retrieval}             & gpt-3.5-turbo  & \ding{56}     & 0.383          & \textbf{0.521} & 0.369          & 0.448          & 0.133          & 0.237          & 0.295          & 0.402          \\
ReAct~\cite{yao_react_2023}                    & gpt-3.5-turbo  & \ding{52}     & 0.317          & 0.411          & 0.312          & 0.387          & 0.136          & 0.220          & 0.255          & 0.339          \\
LPKG~\cite{wang_learning_2024}                 & gpt-3.5-turbo  & \ding{52}     & 0.364          & 0.510          & 0.379          & 0.452          & 0.142          & 0.236          & 0.295          & 0.399          \\
\midrule
\textcolor{gray!85}{NoCoT~\cite{deepseekai2025deepseekr1incentivizingreasoningcapability}} & \textcolor{gray!85}{DeepSeek-R1} & \textcolor{gray!85}{\ding{56}} & \textcolor{gray!85}{0.384} & \textcolor{gray!85}{0.515} & \textcolor{gray!85}{0.442} & \textcolor{gray!85}{0.534} & \textcolor{gray!85}{0.143} & \textcolor{gray!85}{0.267} & \textcolor{gray!85}{0.323} & \textcolor{gray!85}{0.439} \\
\midrule
\textbf{ORACLE}            & gpt-3.5-turbo  & \ding{52}     & \textbf{0.396} & 0.518          & \textbf{0.468} & \textbf{0.547} & \textbf{0.156} & \textbf{0.242} & \textbf{0.340} & \textbf{0.436} \\
\bottomrule
\end{tabular}
\caption{Main results on multi-hop QA benchmarks. Our proposed method, ORACLE, is compared against baselines on HotPotQA, 2WikiMQA, and Musique using GPT-3.5-Turbo. The ``Planning'' column indicates whether a method explicitly plans its reasoning steps (\ding{52}) or not (\ding{56}). \textbf{Bold} values mark the best performance, excluding the reference row ($^*$). The DeepSeek-R1 row, shown in gray text, is for contextual reference only and is not part of the primary comparison.}
\label{tab:main_results}
\end{table*}

\section{Experiments}
In this section, we evaluate the performance of our method on three datasets. Furthermore, we analyze the characteristics of MQA problems and conduct detailed analysis to demonstrate the effectiveness of our approach.
\subsection{Experimental Setup}
\subsubsection{Datasets.}
We conducted experiments on the following three MQA datasets: HotPotQA~\cite{yang2018hotpotqa}, 2WikiMQA~\cite{xanh2020_2wikimultihop}, and MuSiQue~\cite{trivedi2022musique}.  Similar to the previous methods~\cite{wang_learning_2024, shao2023enhancing}, we sampled 500 questions from the development set of each dataset. Specifically, for HotPotQA, we randomly sampled 500 questions from the LPKG~\cite{wang_learning_2024} subset. For 2WikiMQA, we randomly sampled 500 questions. For MuSiQue, we selected 500 questions in a 2:2:1 ratio based on 2-hop (2p), 3-hop (3p), and 4-hop (4p) questions.

\subsubsection{Baselines.}

We compare our framework to various baselines: 
\begin{itemize}
    \item \textbf{NoCoT}: The LLM is instructed to directly answer the input question without additional reasoning.
    \item \textbf{CoT}: Following Chain-of-Thought~\citep{wei_chain--thought_2023}, we instruct the LLM to ``Think step by step" before providing the final answer.
    \item \textbf{RAG}: The prompt sent to the LLM includes both the original question and retrieved information related to it.
    \item \textbf{ReAct}: The ReAct approach~\cite{yao_react_2023} guides an LLM to solve problems by cyclically generating CoTs along with an action using external tools. The results of prior cycles are utilized in the next cycle.
    \item \textbf{LPKG}: This method creates code-formatted planning demonstrations by verbalizing logical patterns from a KG~\cite{wang_learning_2024}. These demonstrations are then used in a prompt to guide an LLM via in-context learning to generate decomposed plans.
\end{itemize}
     
\subsubsection{Implementation Details.}
Unless specified otherwise, all experiments utilized the \textit{gpt-3.5-turbo} API\footnote{Developed by OpenAI; the specific model version we use is gpt-3.5-turbo-0125.} as the base LLM. For non-retrieval baselines (e.g., NoCoT, CoT, and RAG), models were prompted to enclose final answers in \texttt{<answer>} tags to standardize evaluation. The NoCoT (DeepSeek-R1) baseline adopted the same prompt structure but used the \textit{DeepSeek-R1} model\footnote{Developed by DeepSeek; the specific model version we use is DeepSeek-R1-0528.}.
For all retrieval-based methods (ReAct, LPKG, and our proposed ORACLE), we implemented a unified retriever module.
The ReAct baseline integrated this retriever with \textit{gpt-3.5-turbo} as its core agent. Similarly, LPKG employed \textit{gpt-3.5-turbo} to perform its in-context learning-based planning. As a method designed for powerful, general-purpose models, LPKG provides a strong point of comparison, which contrasts with our training-free framework. 
Detailed implementation can be found in the Appendix.

\subsubsection{Evaluation Metrics}
We used Exact Match (EM) and F1 Score as evaluation metrics across all MQA datasets. Both metrics first apply a normalization step that ensures a fair and case-insensitive comparison.
\begin{itemize}
    \item \textbf{EM Score}: This is a strict, all-or-nothing metric. It awards a score of 1 only if the normalized prediction string is identical to the normalized ground truth string, and 0 otherwise.
    \item \textbf{F1 Score}: This metric provides a more flexible, token-level evaluation of performance. It treats the prediction and ground truth as bags of words (tokens). The F1 score gives partial credit for answers that have overlapping words with the ground truth, making it a valuable measure for questions where answers can be phrased in slightly different ways.
\end{itemize}

\subsection{Experiment Results}

The experimental results, presented in Table \ref{tab:main_results}, show that our proposed method, \textsc{ORACLE}, achieves state-of-the-art or highly competitive performance across the HotPotQA, 2WikiMQA, and MuSiQue datasets. Overall, \textsc{ORACLE} secures the highest average EM score of 0.340, demonstrating its robust reasoning capabilities.

Specifically, on HotPotQA, \textsc{ORACLE} achieves the highest EM score of 0.396, although the RAG baseline obtains a slightly higher F1 score of 0.521. The performance advantage of our method is most pronounced on 2WikiMQA, where \textsc{ORACLE} establishes a new state-of-the-art with an EM of 0.468 and an F1 score of 0.547. On the challenging MuSiQue dataset, our approach again secures the top EM score of 0.156, while the reference NoCoT with DeepSeek-R1 achieves the highest F1 score of 0.267.

We highlight several key findings considering these aforementioned experimental results. First, the inferior performance of both NoCoT and CoT is anticipated, as these methods lack access to the external knowledge essential for MQA. Notably, CoT often underperforms NoCoT. A plausible explanation is that for fact-intensive MQA, compelling the model to generate reasoning steps without sufficient evidentiary grounding can introduce hallucinations or logical fallacies, thereby degrading performance. We provide case studies in the appendix.

Second, \textsc{ORACLE} surpasses strong baselines like ReAct and LPKG. ReAct's iterative, agent-based process is susceptible to error propagation, where an early mistake in retrieval or reasoning can compound and derail the entire process. Meanwhile, LPKG, which relies on in-context examples for planning, employs a strategy that can be too rigid to generalize across the diverse reasoning paths required by complex MQA. In contrast, the promising performance of \textsc{ORACLE}, particularly in achieving the highest EM score on every dataset, suggests that its ontology-driven planning and decomposition strategy is more flexible and resilient to the accumulation of errors.

\section{Analysis}
\label{sec:analysis}
In this section, we conduct experiments to analyze the contributions of different components in our approach and further explore the factors influencing complex reasoning problems involving commonsense.

\subsection{What is the contribution of each component?}

\begin{table}[t]
\centering
\renewcommand{\arraystretch}{1.2}

\begin{tabular}{l S[table-format=1.4] S[table-format=1.4]}
\toprule
& {EM} & {F1} \\
\midrule
w/o Ontology & 0.338 & 0.424 \\
w/o FOL   & 0.304 & 0.408 \\
ORACLE     & 0.396   & 0.518   \\
\bottomrule
\end{tabular}
\caption{Ablation study results on the HotPotQA dataset.}
\label{tab:ablation_hotpotqa_lines}
\end{table}

To ascertain the contribution of our method's core components, we conduct an ablation study on the HotPotQA dataset. We individually remove two key modules: (1) Entity and Relation Extraction (corresponding to \texttt{w/o Ontology}) and (2) Logical Analysis (corresponding to \texttt{w/o FOL}). Table~\ref{tab:ablation_hotpotqa_lines} shows the results, where \textsc{ORACLE} represents our full model.

\textbf{Ablating the Logical Analysis Module (w/o FOL)} results in performance degradation. The $EM$ score drops sharply from \textbf{0.396} to \textbf{0.304}, a relative decrease of approximately \textbf{23.2\%}. The $F1$ score shows a similarly steep decline from \textbf{0.518} to \textbf{0.408}. This is because, without a structured reasoning plan from this module, the model is prone to logical missteps. This outcome underscores that an explicit logical plan is critical for correctly connecting reasoning steps to derive the final answer.

\textbf{Ablating the Entity and Relation Extraction Module (w/o Ontology)} also causes a considerable performance drop, with the $EM$ score falling from \textbf{0.396} to \textbf{0.338}. This indicates that explicitly identifying key entities and their relations provides essential grounding for the reasoning process. Without these anchors, the model is more susceptible to hallucinating connections or failing to identify the crucial bridge information, thus compromising the integrity of the reasoning chain.

\subsection{Fine-grained Comparison}

\begin{figure}[ht]
    \centering
    \includegraphics[width=0.95\linewidth]{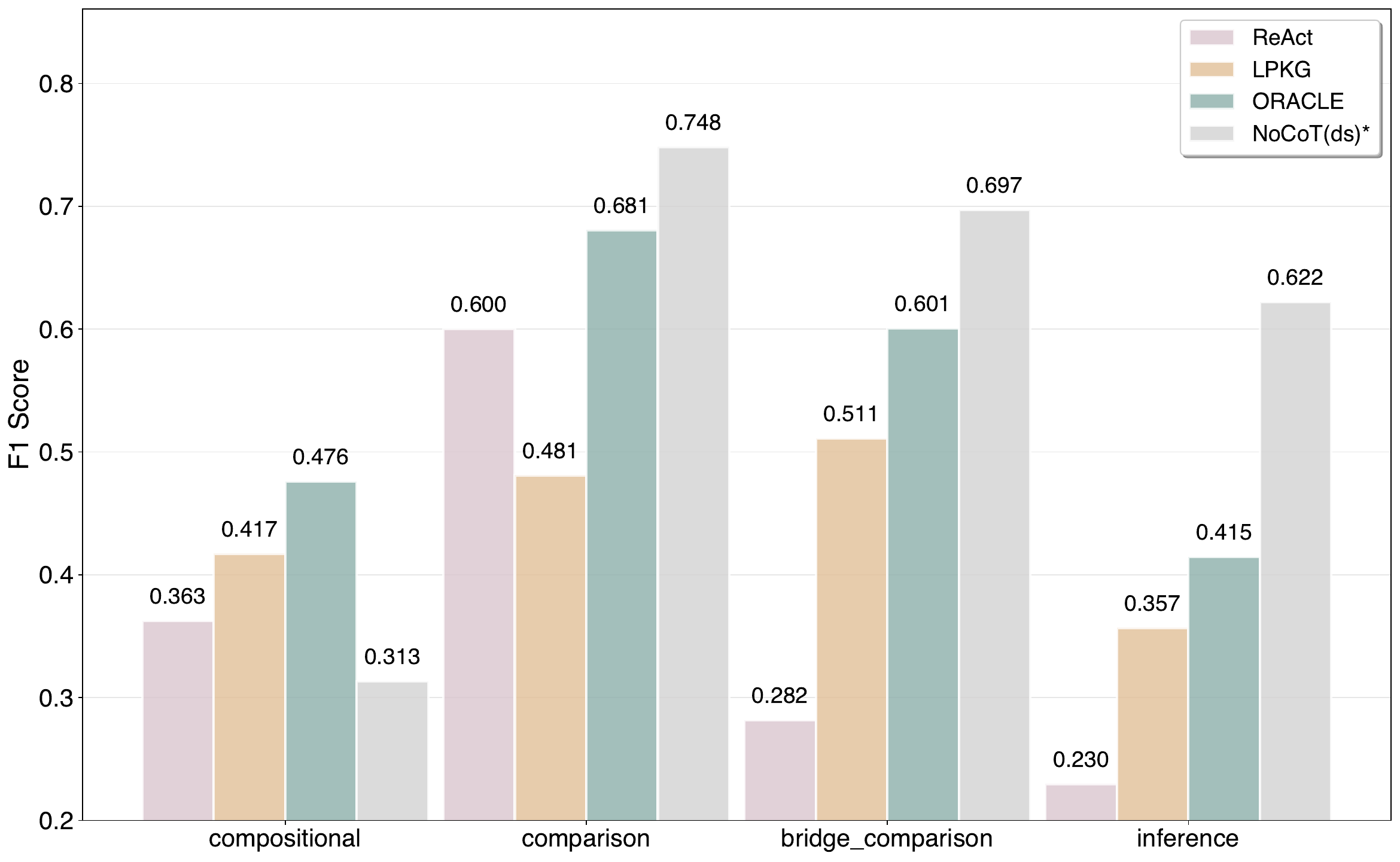}
    \caption{Fine-Grained Performance Analysis on 2WikiMQA by Reasoning Type. The Metric Shown Is F1 Score. NoCoT (ds)$^*$ Is Included for Reference Only, Where ``ds'' Denotes DeepSeek-R1.}
    \label{fig:2wikimqa_fine_grain}
\end{figure}

To pinpoint the specific advantages of our method, we conduct a fine-grained performance analysis on the 2WikiMQA dataset, categorizing questions into four types. The results are shown in Figure~\ref{fig:2wikimqa_fine_grain}.

Across all categories, our method, \textsc{ORACLE}, consistently outperforms the other planning-based methods, ReAct and LPKG. The advantage is observed on \textit{Compositional} questions, where ORACLE achieves the highest F1 score (0.476), underscoring the effectiveness of its planning and decomposition strategy. For more complex reasoning tasks such as \textit{Comparison}, \textit{Bridge-Comparison}, and \textit{Inference}, ORACLE maintains a strong lead over other planning methods. While the NoCoT (ds) reference indicates the upper-bound performance on MQA tasks, our method's consistent top-ranking performance among planning-based approaches highlights its robustness and superior reasoning capabilities across a diverse set of challenges.

\subsection{Impact of Reasoning Path Quality on Performance}

\begin{figure}[ht]
    \centering
    \includegraphics[width=0.95\linewidth]{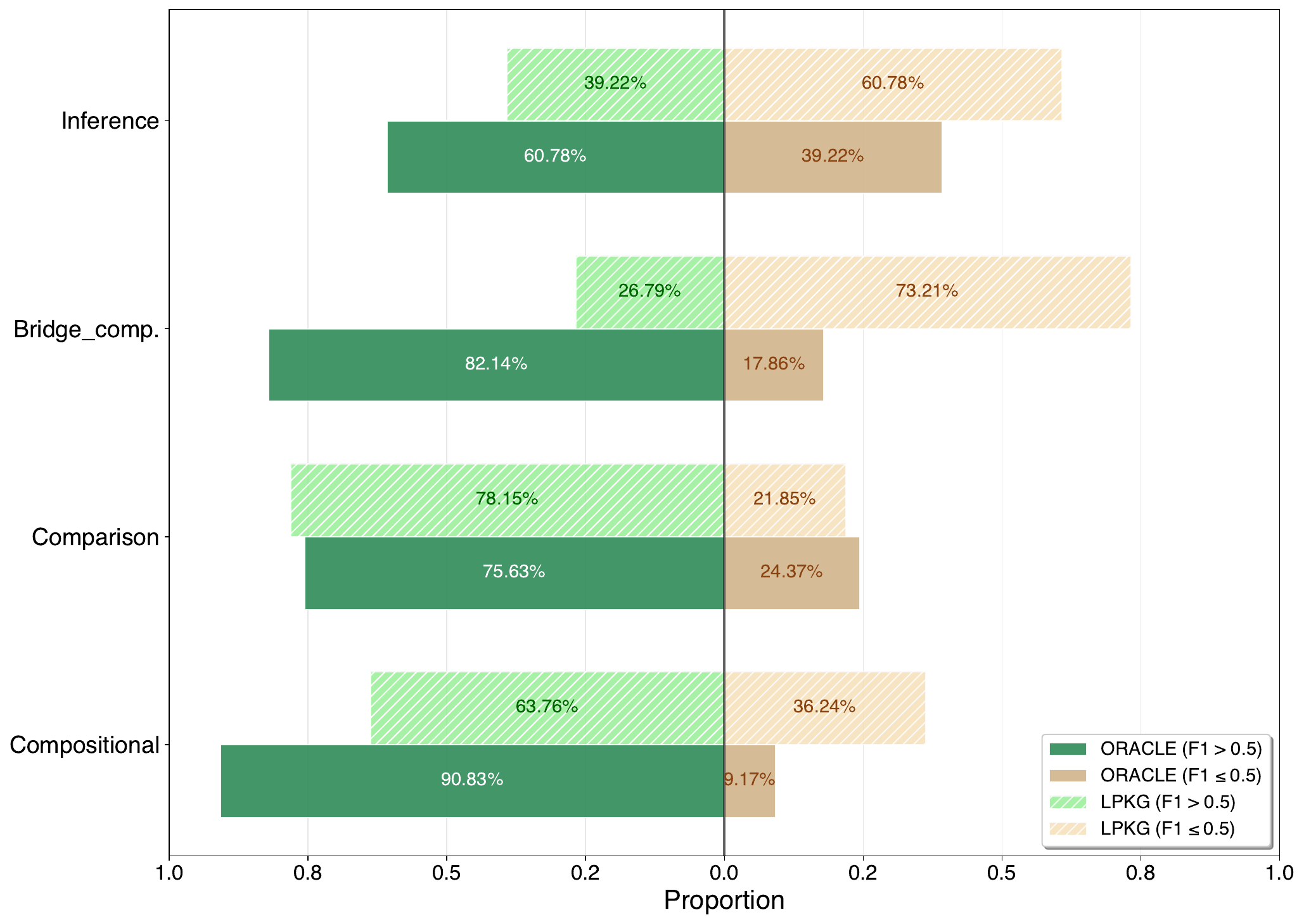}
    \caption{F1 Score Distribution for LPKG and ORACLE Methods. The Bars Illustrate the Percentage of Questions for Which Each Method Attained a High ($F_1 > 0.5$) or Low ($F_1 \le 0.5$) F1 Score.}
    \label{fig:reasoning_f1_distribution} 
\end{figure}

\begin{figure}[ht]
    \centering
    \includegraphics[width=0.95\linewidth]{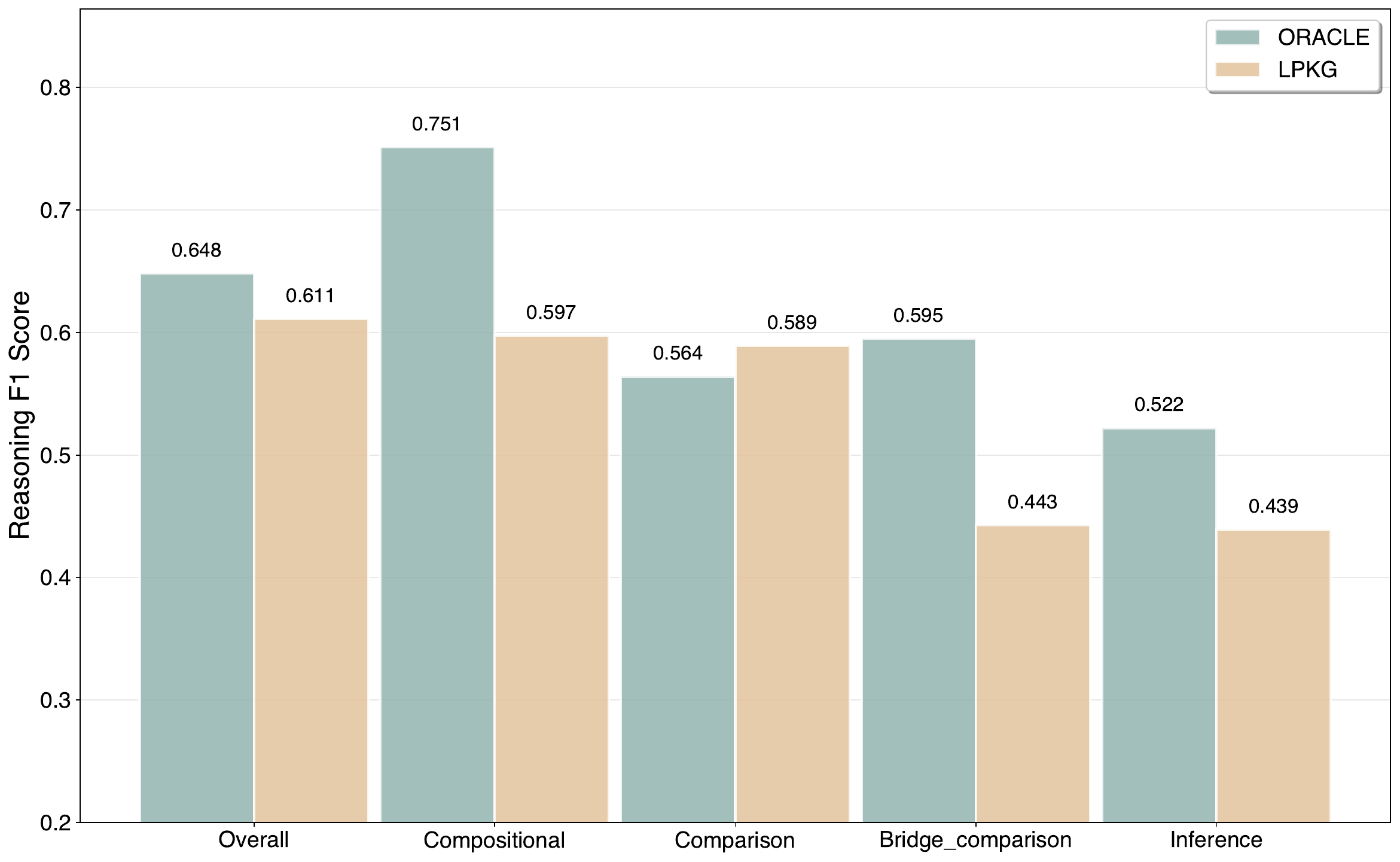}
    \caption{Comparison of Reasoning F1 Scores Between LPKG and ORACLE.}
    \label{fig:reasoning_f1_comparison}
\end{figure}

To quantitatively assess the quality of the generated reasoning process, we define a \textbf{Reasoning $F_1$ score} (see the Appendix for detailed implementation). This metric evaluates the lexical overlap between the model-generated reasoning path and the ground-truth evidence chain. A higher score signifies a reasoning process that is more aligned with the gold standard logic. To understand its impact on final answer accuracy, we segment the analysis based on whether the reasoning path is of high quality (\textbf{Reasoning $F_1 > 0.5$}) or low quality (\textbf{Reasoning $F_1 \le 0.5$}).

Our analysis reveals two critical findings. First, \textbf{ORACLE consistently produces higher-quality reasoning paths}. As shown in Figure~\ref{fig:reasoning_f1_comparison}, ORACLE achieves a higher average Reasoning $F_1$ score than LPKG across all question types, with an overall score of 0.648 compared to LPKG's 0.611. This indicates that ORACLE's planning module is fundamentally more effective. Figure~\ref{fig:reasoning_f1_distribution} reinforces this by showing the respective proportion of high- and low-quality paths. For every question type, ORACLE generates a larger fraction of high-quality paths. For example, on \textit{Compositional} questions, 90.83\% of ORACLE's reasoning paths exceed the $F_1 > 0.5$ threshold, whereas only 63.76\% of LPKG's paths do.

\begin{figure*}[ht]
    \centering
    \includegraphics[width=0.95\linewidth]{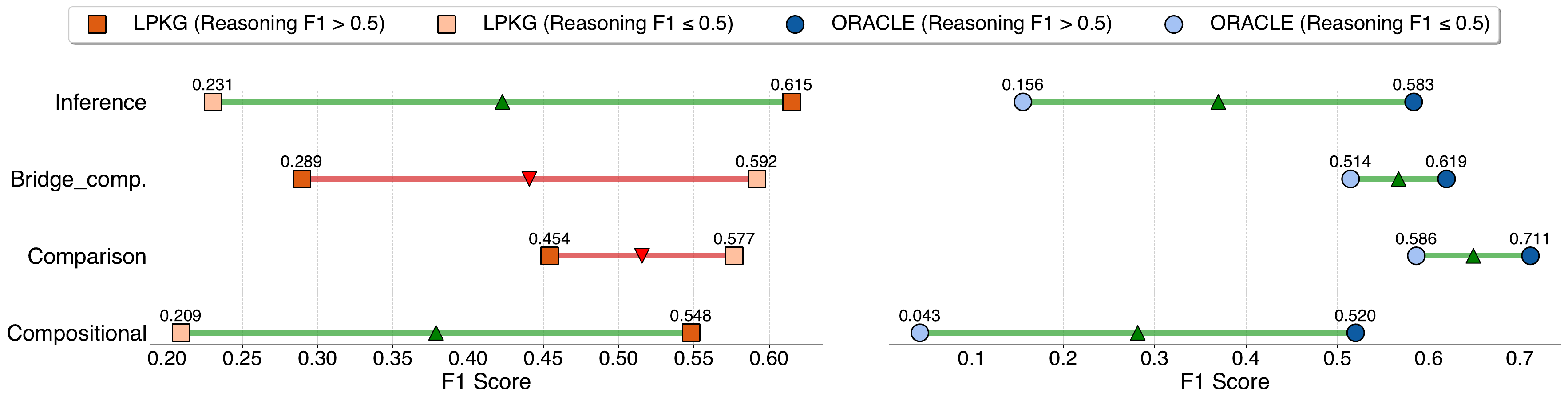}
    \caption{Impact of Intermediate Reasoning Quality. The Analysis Is Presented for Both the LPKG and ORACLE Methods Across the Four Reasoning Categories. The Line Color Indicates the Correlation Between Reasoning Quality and Final Answer Quality for That Category: Green Indicates a Positive Correlation, While Red Indicates a Weaker or Negative Correlation.}
    \label{fig:reasoning_correlation_analysis} 
\end{figure*}

Second, \textbf{ORACLE exhibits much greater faithfulness to its reasoning path}, a crucial factor for reliable and interpretable AI. Figure~\ref{fig:reasoning_correlation_analysis} illustrates this dynamic clearly.
\begin{itemize}
    \item For \textbf{ORACLE}, high-quality reasoning (Reasoning $F_1 > 0.5$, dark blue circles) consistently leads to higher final F1 scores than low-quality reasoning (Reasoning $F_1 \le 0.5$, light blue circles). This demonstrates that its final answer is a direct and reliable consequence of its explicit reasoning process.
    \item In contrast, \textbf{LPKG} often achieves high performance despite a flawed reasoning path. For \textit{Bridge\_comp} and \textit{Inference} questions, LPKG's final F1 score is paradoxically higher when its reasoning is flawed. For instance, on \textit{Inference} tasks, LPKG scores 0.615 with low-quality reasoning paths but only 0.231 with high-quality ones.
\end{itemize}
This behavior suggests that LPKG frequently disregards its generated plan and instead relies on the parametric knowledge of the base LLM to find an answer. While this may occasionally lead to a correct result, it reveals a critical flaw: the model's reasoning is not a reliable indicator of its final output, making it less trustworthy and harder to debug. In contrast, ORACLE's strong performance is directly attributable to its superior and more faithful reasoning capabilities.

\subsection{Impact of Subproblem Count on Performance}

\begin{figure}[ht]
    \centering
    \includegraphics[width=0.95\linewidth]{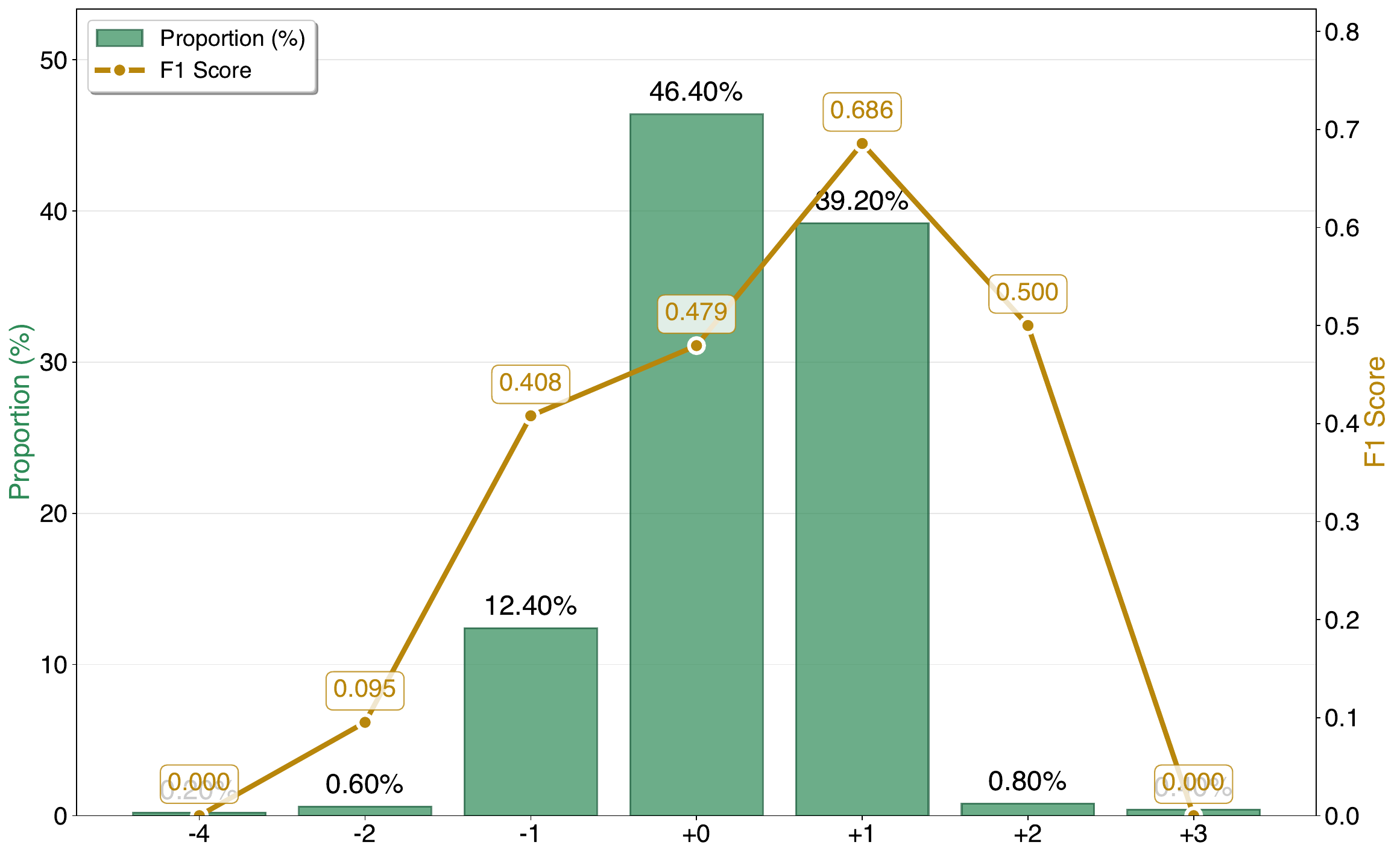}
    \caption{Analysis of Subproblem Count Deviation and Its Impact on F1 Score (ORACLE Method). The Bar Chart (Left Axis) Displays the Proportion of Questions for Each Deviation Value, Where ``+0'' Indicates the Correct Number of Subproblems. The Line Chart (Right Axis) Illustrates the Corresponding Impact of This Deviation on the Final F1 Score.}
    \label{fig:subproblem_deviation_analysis} 
\end{figure}

To analyze the relationship between the number of decomposed sub-questions and final answer accuracy, we first establish a ground-truth step count for each question using the evidence triplets from the 2WikiMQA dataset (see the Appendix for detailed implementation). We define \textit{deviation} as the difference between the number of sub-questions our method generates and this ground-truth count. 

As illustrated in Figure~\ref{fig:subproblem_deviation_analysis}, our method's decomposition aligns closely with the ground truth. A perfect match ($+0$ deviation) is the most frequent outcome, occurring in 46.40\% of cases. Overall, 98.00\% of generated plans deviate by at most one step ($\pm1$), with 39.20\% over-decomposing by one ($+1$) and 12.40\% under-decomposing by one ($-1$).

Notably, the model's performance, measured by the F1 score, peaks at \textbf{0.686} for a $+1$ deviation. This score is substantially higher than the F1 score of \textbf{0.479} achieved for a perfect match ($+0$ deviation). This suggests a clear asymmetry in reasoning errors: including a redundant step is far less detrimental, and often beneficial, than omitting a crucial one. Conversely, under-decomposition ($-1$ deviation) causes a drop in performance to an F1 score of \textbf{0.408}, indicating that failing to generate a required sub-question impairs the reasoning process.

\subsection{More experiments across Model Sizes on Qwen2.5}

Firstly, we note that we did not extend validation to dedicated reasoning models, since without fine-tuning, such models typically struggle with complex instruction-following. Thus, our experiments focus on general-purpose models where ontology-guided reasoning can be more faithfully evaluated.

To further validate the effectiveness of our method beyond the main setting, we evaluate LPKG and ORACLE on a series of models, including Qwen2.5 with 7B, 14B, 32B, and 72B parameters. The results are summarized in Table~\ref{tab:cross-model-vertical}, covering three metrics: Exact Match (EM), F1, and Reasoning F1. 

\begin{table}[t]
\centering
\begin{tabular}{lcccc}
\toprule
\textbf{Model} & \textbf{Method} & \textbf{EM} & \textbf{F1} & \textbf{Reason F1} \\
\midrule
\multirow{2}{*}{GPT-3.5} 
 & LPKG   & 0.379 & 0.452 & 0.611 \\
 & ORACLE & \textbf{0.469} & \textbf{0.548} & \textbf{0.650} \\
\midrule
\multirow{2}{*}{Qwen2.5-7B} 
 & LPKG   & 0.432 & 0.533 & \textbf{0.620} \\
 & ORACLE & 0.447 & 0.507 & 0.598$^{\star}$ \\
\midrule
\multirow{2}{*}{Qwen2.5-14B} 
 & LPKG   & 0.471 & 0.571 & 0.637 \\
 & ORACLE & \textbf{0.494} & \textbf{0.587} & \textbf{0.654} \\
\midrule
\multirow{2}{*}{Qwen2.5-32B} 
 & LPKG   & 0.440 & 0.531 & 0.665 \\
 & ORACLE & \textbf{0.527} & \textbf{0.635} & \textbf{0.679} \\
\midrule
\multirow{2}{*}{Qwen2.5-72B} 
 & LPKG   & 0.485 & 0.584 & 0.659 \\
 & ORACLE & \textbf{0.557} & \textbf{0.638} & \textbf{0.688} \\
\bottomrule
\end{tabular}
\caption{Vertical Comparison of LPKG and ORACLE Across Different Models. Metrics Include Exact Match (EM), F1, and Reasoning F1. The Superscript $\star$ Indicates the Only Case Where LPKG Outperforms ORACLE in Reasoning F1.}
\label{tab:cross-model-vertical}
\end{table}

From the results, we observe the following:

\begin{itemize}
    \item \textbf{Overall Superiority of ORACLE.} Except for the 7B model, ORACLE consistently outperforms LPKG across all metrics and model sizes. This validates that our ontology-driven approach yields stable improvements with larger and more capable models. 
    
    \item \textbf{7B Exception.} On Qwen2.5-7B, LPKG achieves higher Reasoning F1 (0.6202 vs.\ 0.5983). We hypothesize that the limited parameter capacity restricts the model's ability to leverage ontology guidance effectively. In contrast, LPKG benefits more from its detailed prompt design, which encodes more external knowledge rather than relying on the model's internal reasoning.
    
    \item \textbf{Scaling Effects.} As parameter size increases (32B vs.\ 72B), F1 scores become similar, but Reasoning F1 continues to improve (0.6792 $\rightarrow$ 0.6879). This suggests that larger models not only strengthen internal knowledge representation but also enhance their ability to integrate information for reasoning.
    
    \item \textbf{Architecture Efficiency.} Comparing GPT-3.5 (175B) with Qwen2.5-14B, we find comparable or even superior performance from the smaller Qwen2.5 model, indicating that newer architectures deliver higher parameter efficiency. This trend supports the view that sparsity and architectural advances are as critical as sheer scale.
\end{itemize}

\section{Conclusion}
This paper introduces the ORACLE framework, designed to enhance the performance of LLMs on MQA tasks. The framework begins by leveraging the concept of an ontology from knowledge representation to guide the LLM in extracting relevant ontological structures from the query. These extracted ontologies are then transformed into FOL formulas, creating a structured reasoning chain to aid in solving the MQA problem. Subsequently, the original question is decomposed based on these ontologies and FOL representations. The effectiveness of the ORACLE method was validated on standard MQA datasets, with experiments demonstrating its superior and more accurate reasoning capabilities. This approach offers a new perspective on factual reasoning for LLMs.

\bibliography{aaai2026}

\clearpage
\appendix

\section{Appendix}
\subsection{Prompt Content}
In this section, we provide the detailed prompts utilized for the different reasoning decomposition methods in our experiments.

The prompt for the \texttt{ReAct} method, as shown in Table~\ref{tab:react_prompt}, guides the model to solve problems through an iterative cycle of Thought, Action, and Observation. This approach allows the model to dynamically reason and interact with an external knowledge source.

Table~\ref{tab:our_prompt} presents the prompt for our proposed method, ORACLE. It employs a more structured, five-step decomposition process that requires the model to first analyze the problem's structure (type identification, entity extraction, logical conversion) before breaking it down into a sequence of simpler sub-questions.

For the LPKG method, we used the prompt as described in the original LPKG.

\begin{table}[htbp]
\small
\centering
\begin{tabular}{@{}>{\RaggedRight\arraybackslash}m{65pt}>{\RaggedRight\arraybackslash}m{0.65\linewidth}@{}}
\toprule
\textbf{Method} & \textbf{Detailed Prompt} \\ 
\midrule
\texttt{ReAct} & 
Solve a question answering task with interleaving Thought, Action, Observation steps. Thought can reason about the current situation, and Action can be three types: \newline
(1) \texttt{Search[entity]}, which searches the exact entity on Wikipedia and returns the first paragraph if it exists. If not, it will return some similar entities to search. \newline
(2) \texttt{Lookup[keyword]}, which returns the next sentence containing the keyword in the current passage. \newline
(3) \texttt{Finish[answer]}, which returns the answer and finishes the task. \newline
\texttt{\{examples\}} \newline
\texttt{Question:}
\\
\bottomrule
\end{tabular}%
\caption{The prompt for the ReAct method.}
\label{tab:react_prompt}
\end{table}

\begin{table}[htbp]
\small
\centering
\begin{tabular}{@{}>{\RaggedRight\arraybackslash}m{65pt}>{\RaggedRight\arraybackslash}m{0.65\linewidth}@{}}
\toprule
\textbf{Method} & \textbf{Detailed Prompt} \\ 
\midrule
\texttt{ORACLE} & 
Your task is to decompose complex reasoning problems into a series of sub-questions. Please follow these steps: \newline
1. \textbf{Problem Type Identification}: Determine the problem type (2p/3p path query, 2i/3i intersection query, or ip/pi hybrid query) \newline
2. \textbf{Entity and Relation Extraction}: List all key entities and their relationships \newline
3. \textbf{Logical Formula Conversion}: Convert the problem into a formal logical expression \newline
4. \textbf{Logical Interpretation}: Explain the meaning of the logical expression in natural language \newline
5. \textbf{Sub-question Decomposition}: Break down the original problem into an ordered sequence of sub-questions, clearly labeling each sub-question's answer variable \newline
\texttt{\{examples\}} \newline
\texttt{\#\#\# Your turn! Please decompose complex reasoning problem.} \newline
\textbf{Question}:
\\
\bottomrule
\end{tabular}%
\caption{The prompt for the ORACLE method.}
\label{tab:our_prompt}
\end{table}

\subsection{Retriever Implementation}
Conventional retriever implementations for MQA typically utilize the complete Wikipedia dump, which includes \emph{psgs\_w100.tsv.gz} and \emph{wikipedia\_embeddings.tar}, as the knowledge base. However, loading the \emph{wikipedia\_embeddings.tar} file requires substantial computational resources, specifically 2x 80G A100 GPUs, making this approach prohibitively expensive.

To address this, we observed that MQA datasets generally provide a context containing the necessary paragraphs to answer the question, along with some distractor paragraphs. Consequently, we propose using this provided context as the search space for the retriever. This method offers two key advantages: it filters out the vast amount of irrelevant information present in the full Wikipedia dump and significantly reduces the demand for computational resources. By treating the provided context as the retrieval corpus, we can mitigate the possibility of the retriever acting as a performance bottleneck, thereby enabling a more accurate evaluation of our proposed method's effectiveness.

In our implementation, the retriever matches the query against the provided context. Let $q$ be the input query and $C = \{p_1, p_2, \dots, p_n\}$ be the set of paragraphs in the given context. The retriever identifies a subset of relevant paragraphs, $C_{\text{retrieved}}$, by calculating a similarity score between the query and each paragraph. A paragraph $p_i$ is returned if its similarity score exceeds a predefined threshold $\tau$. This process can be formally expressed as:
$$
C_{\text{retrieved}} = \{p_i \in C \mid \text{sim}(q, p_i) \ge \tau\}
$$
where $\text{sim}(q, p_i)$ is the similarity function.
For the RAG baseline, we simplify this process by retrieving the entire context, including both golden and distractor paragraphs. 
For a detailed implementation, please refer to the accompanying code.

\subsection{Reasoning Path and Subquestion Count Implementation}
Given the scarcity of research evaluating the internal reasoning processes of LLMs, and substantial evidence showing that models can produce correct answers via fallacious reasoning, a more granular assessment is necessary. As noted by \citet{ju-etal-2024-investigating}, when faced with a query such as, ``On which continent is the home country of the 2022 FIFA World Cup winner?'' an LLM might not follow the correct reasoning chain, ``Winner $\rightarrow$ Argentina $\rightarrow$ South America.'' Instead, it may leverage a shortcut learned during training, directly linking ``2022 World Cup Winner'' to ``South America.''

\begin{table*}[!h]
\centering
\begin{tabular}{l p{6cm} p{6cm}}
\toprule
& \multicolumn{1}{c}{\textbf{NoCoT}} & \multicolumn{1}{c}{\textbf{CoT}} \\
\midrule
\textbf{Question} & \multicolumn{2}{c}{What nationality is Prince Napoléon Bonaparte's father?} \\
\midrule
\textbf{Prompt} & You will be provided with a task description, a question and an answer format.
You should only respond with the final answer in the format ``Answer: ". & You will be provided with a task description, a question and an answer format.
You should think step by step.
In the end, put the final answer in the format ``Answer: ". \\
\midrule
\textbf{Model Output} & Answer: French. & The model generates a generic reasoning plan:
\begin{enumerate}
    \item Find information about Prince Napoléon Bonaparte's father.
    \item Check his father's name and background.
    \item Look for his father's nationality.
\end{enumerate} Answer: Italian.\\

\midrule
\textbf{Final Answer} & French & Italian \\
\midrule
\textbf{Result} & \textcolor{teal}{\textbf{Correct}} & \textcolor{red}{\textbf{Incorrect}} \\
\bottomrule
\end{tabular}
\caption{Comparison of NoCoT and CoT}
\label{tab:case_study}
\end{table*}

Consequently, we propose an evaluation framework to compare the LPKG and ORACLE methods from the perspective of their reasoning processes. Our implementation leverages the 2WikiMQA dataset, which includes a \texttt{question\_decomposition} field. This field serves as our ground truth, providing a list of interconnected, single-hop sub-questions (formatted as \texttt{(head entity, relation, tail entity)} triples) that constitute the correct reasoning path. We consolidate this sequence of triples into a single string to represent the ground-truth path.

For our evaluation, we transform the decomposed sub-questions and their answers from both ORACLE and LPKG into the same concatenated string of triples. 
To ensure fair comparison with the gold reasoning path, we first apply a preprocessing step to the generated reasoning text, including \emph{lowercasing}, \emph{stopword removal}, and \emph{lemmatization}. 
This yields a normalized representation of the reasoning process that contains only the essential \emph{entities} and \emph{relations}. 
We then align this representation with the ground-truth reasoning path provided in the dataset, also expressed as a concatenated sequence of triples. 

Formally, let $G = \{g_1, g_2, \dots, g_m\}$ denote the set of tokens (entities and relations) in the ground-truth reasoning path, and 
$P_{\text{gen}} = \{p_1, p_2, \dots, p_n\}$ denote the tokens extracted from the model-generated reasoning path after preprocessing. 
We compute precision $P$ and recall $R$ as:  
\begin{equation*}
P = \frac{|G \cap P_{\text{gen}}|}{|P_{\text{gen}}|}, 
\quad 
R = \frac{|G \cap P_{\text{gen}}|}{|G|}
\end{equation*}
and define the \textbf{Reasoning F1} score as:  
\begin{equation*}
\text{ReasonF1} = \frac{2 \cdot P \cdot R}{P + R}.
\end{equation*}

In parallel, we also compute the \emph{sub-question count deviation} to measure the granularity of decomposition. 
Let $k^{*}$ denote the ground-truth number of reasoning steps and $k_{\text{gen}}$ the number generated by the model. 
The deviation is defined as: 
\begin{equation*}
\Delta k = k_{\text{gen}} - k^{*}.
\end{equation*}
This enables us to analyze how over-decomposition or under-decomposition impacts task performance. 

\subsection{Case Studies}\label{appendix:case-studies}
\subsubsection{CoT and NoCoT}


To provide a concrete illustration of our findings, we present a case study that highlights the potential drawbacks of CoT, particularly for models with more limited reasoning capabilities.

The results, detailed in Table~\ref{tab:case_study}, show a clear performance degradation when CoT is applied.

As shown in Table~\ref{tab:case_study}, the direct answering approach correctly identifies the nationality as \textbf{French}. The model effectively uses its trained knowledge to retrieve the fact directly.

In contrast, the CoT method incorrectly answers \textbf{Italian}. We hypothesize that for less capable models, CoT is counterproductive on knowledge-intensive questions because the self-generated reasoning plan flattens the final answer's logit distribution. This diffusion of attention away from the key entity and across abstract reasoning steps causes the model to retrieve a related but incorrect fact.

\subsubsection{LPKG and ORACLE}
\label{sec:case-study-onto}
To further illustrate the advantages of our ontology-driven decomposition, 
we present two representative case studies comparing ORACLE with the LPKG baseline.

\begin{table*}[!h]
\centering
\begin{tabular}{l p{6.5cm} p{6.5cm}}
\toprule
& \multicolumn{1}{c}{\textbf{LPKG}} & \multicolumn{1}{c}{\textbf{ORACLE}} \\
\midrule
\textbf{Question} & \multicolumn{2}{c}{Which film has the director who died first, \emph{The Piper's Price} or \emph{The Blonde From Singapore}?} \\
\midrule
\textbf{Reasoning Process} 
& \begin{itemize}
    \item Which film is directed by the director of \emph{The Piper}? $\rightarrow$ \emph{The Faded Woman}
    \item Which film is directed by the director of \emph{The Blonde From Singapore}? $\rightarrow$ \emph{The Blonde from Peking}
\end{itemize} 
& \begin{itemize}
    \item Who directed \emph{The Piper's Price}? $\rightarrow$ Joe De Grasse
    \item Who directed \emph{The Blonde From Singapore}? $\rightarrow$ Edward Dmytryk
    \item When did [Joe De Grasse] die? $\rightarrow$ May 25, 1940
    \item When did [Edward Dmytryk] die? $\rightarrow$ July 1, 1999
    \item Compare death dates $\rightarrow$ May 25, 1940
\end{itemize} \\
\midrule
\textbf{Ground Truth Path} & \multicolumn{2}{p{13cm}}{
The Piper's Price $\rightarrow$ Joe De Grasse $\rightarrow$ May 25, 1940; 
The Blonde From Singapore $\rightarrow$ Edward Dmytryk $\rightarrow$ July 1, 1999} \\
\midrule
\textbf{ReasonF1} & 0.30 & 0.66 \\
\bottomrule
\end{tabular}
\caption{Case study comparing LPKG and ORACLE on a film director comparison question.}
\label{tab:case_study_piper}
\end{table*}

The first case (Table~\ref{tab:case_study_piper}) is a comparison-type query 
asking which film’s director died first. LPKG mistakenly interprets the sub-question 
as “find another film directed by the same director,” which leads to irrelevant reasoning paths. 
In contrast, ORACLE first extracts the ontology to anchor the key entities (films, directors, and death dates), 
then generates a reasoning chain that matches the ground truth. 
This results in a substantially higher ReasonF1, indicating closer alignment with the gold reasoning path. 

\begin{table*}[!h]
\centering
\begin{tabular}{l p{6.5cm} p{6.5cm}}
\toprule
& \multicolumn{1}{c}{\textbf{LPKG}} & \multicolumn{1}{c}{\textbf{ORACLE}} \\
\midrule
\textbf{Question} & \multicolumn{2}{c}{Where was the father of Marianus V of Arborea born?} \\
\midrule
\textbf{Reasoning Process} 
& \begin{itemize}
    \item Who is Marianus V of Arborea? $\rightarrow$ Marianus V was the Judge of Arborea
    \item Where was the father of ``Marianus V was the Judge of Arborea'' born? $\rightarrow$ Castel Genovese
\end{itemize} 
& \begin{itemize}
    \item Who is the father of Marianus V of Arborea? $\rightarrow$ Brancaleone Doria
    \item Where was [Brancaleone Doria] born? $\rightarrow$ Republic of Genoa
\end{itemize} \\
\midrule
\textbf{Ground Truth Path} & \multicolumn{2}{p{13cm}}{
Marianus V of Arborea $\rightarrow$ father $\rightarrow$ Brancaleone Doria $\rightarrow$ place of birth $\rightarrow$ Sardinia} \\
\midrule
\textbf{ReasonF1} & 0.27 & 0.70 \\
\bottomrule
\end{tabular}
\caption{Case study comparing LPKG and ORACLE on a familial relation query.}
\label{tab:case_study_marianus}
\end{table*}

The second case (Table~\ref{tab:case_study_marianus}) is a relation-based query 
about the birthplace of Marianus V of Arborea’s father. 
Here, LPKG drifts to a descriptive path focusing on Marianus V himself, while ORACLE correctly identifies the parent entity (Brancaleone Doria) and queries his birthplace. 
Although minor deviations from the ground answer exist, the ontology-driven reasoning still yields a reasoning chain much closer to the ground truth, again reflected in a higher ReasonF1 score. 

Together, these examples demonstrate that ORACLE not only avoids semantic drift in sub-question decomposition but also consistently produces reasoning paths that are more faithful to the intended logical structure, leading to more interpretable and reliable answers across diverse query types. 
Furthermore, we observe that across multiple runs, ORACLE tends to generate reasoning processes that are relatively stable, a property largely attributable to the guidance of ontology extraction at the planning stage.

\end{document}